\documentclass{article}
\pdfoutput=1
\usepackage{amsmath, amssymb, amsfonts, amsthm, multirow, epsfig,floatrow, graphicx,mathtools, natbib, bbm, pdflscape, verbatim, enumerate, colortbl, booktabs,array, url, xcolor, authblk}
\usepackage{graphicx, caption, subcaption}
\usepackage{hyperref}
\hypersetup{colorlinks=true, linkcolor=blue, citecolor=blue}
\usepackage[utf8]{inputenc}
\usepackage[normalem]{ulem}
\usepackage[ruled,vlined]{algorithm2e}

\def\R{\mathbb{R}}

\def\E{\mathbb{E}}
\def\holder{H\"{o}lder }

\def\H{\mathcal{H}}
\def\A{\mathcal{A}}

\def\O{{O}}

\def\D{\mathcal{D}}

\def\P{\mathbb{P}}

\def\t{\textbf{t}}

\def\lpr{\hat f_{\rm lpr}}

\def\bmax{\beta_{\max}}

\def\0{\textbf{0}}

\newtheorem{proposition}{Proposition}
\newtheorem{lemma}{Lemma}
\newtheorem{definition}{Definition}
\newtheorem{theorem}{Theorem}

\newtheorem{remark}{Remark}

\newcommand{\beq}{\begin{equation}}
\newcommand{\eeq}{\end{equation}}
\newcommand{\beas}{\begin{eqnarray*}}
\newcommand{\eeas}{\end{eqnarray*}}
\newcommand{\bea}{\begin{eqnarray}}
\newcommand{\eea}{\end{eqnarray}}
\newcommand{\bei}{\begin{itemize}}
\newcommand{\eei}{\end{itemize}}
\newcommand{\ben}{\begin{enumerate}}
\newcommand{\een}{\end{enumerate}}
\newcommand{\bet}{\begin{theorem}}
\newcommand{\eet}{\end{theorem}}
\newcommand{\bel}{\begin{lemma}} 
\newcommand{\eel}{\end{lemma}}
\newcommand{\bep}{\begin{proposition}}
\newcommand{\eep}{\end{proposition}}
 \newcommand{\argmin}{\mathop{\rm arg\min}}

\begin{document} 
	
	
\title{Transfer Learning for Nonparametric Regression: Non-asymptotic Minimax Analysis and Adaptive Procedure\footnote{
The research was supported in part by NSF Grant DMS-2015259 and NIH grant
R01-GM129781.
MSC 2010 subject classifications: Primary 62G08; secondary 62L12
Keywords and phrases: Adaptivity, nonparametric regression, optimal rate of convergence, transfer learning
}}  
\author[]{T. Tony Cai and Hongming Pu\footnote{These authors contributed equally to this work.}}
\affil{University of Pennsylvania}
\date{}

\maketitle
	
  
\begin{abstract}
Transfer learning for nonparametric regression is considered. We first study the non-asymptotic minimax risk for this problem and develop a novel estimator called the confidence thresholding estimator, which is shown to achieve the minimax optimal risk up to a logarithmic factor. Our results demonstrate two unique phenomena in transfer learning: auto-smoothing and super-acceleration, which differentiate it from nonparametric regression in a traditional setting. We then propose a data-driven algorithm that adaptively achieves the minimax risk up to a logarithmic factor across a wide range of parameter spaces. Simulation studies are conducted to evaluate the  numerical performance of the adaptive transfer learning algorithm, and a real-world example is provided to demonstrate the benefits of the proposed method.
\end{abstract}




\section{Introduction}

Transfer learning, a technique that utilizes knowledge gained from related source domains to improve performance in a target domain, has gained widespread popularity in machine learning due to its successes across a range of applications, including natural language processing (\citealp{daume2009frustratingly}), computer vision (\citealp{tzeng2017adversarial}), and epidemiology (\citealp{apostolopoulos2020covid}). See the recent survey papers on transfer learning \citep{weiss2016survey,zhuang2020comprehensive} for more examples and in-depth discussions.

Transfer learning has received significant recent attention in statistics, due to its empirical successes. It has been studied in a decision-theoretical framework for a variety of supervised learning problems, such as classification \citep{cai2021transfer, reeve2021adaptive}, high-dimensional linear regression \citep{li2020transfer}, and generalized linear models  \citep{Li2021Transfer-GLM}, as well as unsupervised learning problems, such as Gaussian graphical models \citep{Li2020Transfer-GMM}. Minimax optimal rates of convergence have been established and data-driven adaptive algorithms have been developed. 



In this paper we consider transfer learning for nonparametric regression. Formally, in the target domain one observes independent and identically distributed (i.i.d.)  samples $(X_{i},Y_{i})\stackrel{iid}{\sim} Q$, $i=1,\dots,n_Q$,  with
$$Y_i = f(X_i)+z_i,$$
where $f(\cdot)$ is an unknown function of interest and  $\{z_1,\dots,z_{n_Q}\}$ are  random noises satisfying $\E (z_i |X_i)=0$. Different from the conventional setting, in transfer learning one also has auxiliary data from the source domains. For ease of presentation, we first focus on the case of a single source domain and discuss the case of multiple source domains later. In the single source domain setting, in addition to the samples from the target domain, one observes i.i.d. samples $(X_{i}',Y_{i}')\stackrel{iid}{\sim} P$, $i=1,\dots,n_{P}$, from the source domain with
$$Y_{i}'= g(X_{i}')+z_{i}',$$
where $\{z_{i}'\}$ are i.i.d. random noise satisfying $\E (z_{i}'|X_{i}')=0$ and $g$ is an unknown function. 

In the context of transfer learning for nonparametric regression, the joint distribution $P$ of  $(X_{i}',Y_{i}')$ from the source domain and the joint distribution $Q$ of $(X_i,Y_i)$ from the target domain are different but related. Two popular settings that have been considered in the literature are covariate shift and posterior drift. In the case of covariate shift, the conditional mean functions from the target and source domains, $f$ and $g$, are the same, but the marginal distributions of the covariates, $X_i$ and $X_i'$, are different \citep{shimodaira2000improving, huang2006correcting, wen2014robust}. On the other hand, the posterior drift model assumes that the mean functions, $f$ and $g$, may be different, but the marginal distributions of the covariates, $X_i$ and $X_i'$, are the same. The posterior drift model is a general framework that can be applied to many practical problems, including robotics control \citep{vijayakumar2002statistical, nguyen2008learning, nguyen2008computed, yeung2009learning, cao2010adaptive} and  air quality prediction (\citealp{mei2014inferring, wang2016nonparametric}).  
 
 In this paper, we focus on the posterior drift model, where we assume that the difference between the mean functions, $f$ and $g$, can be well approximated by a polynomial function of a given order in $L_1$ distance. This distance is referred to as the \emph{bias strength}, denoted by $\epsilon$. It controls the similarity between $f$ and $g$ up to a polynomial of a given order. The smaller the bias strength, the more similar $f$ and $g$ are, and vice versa. When the bias strength is zero, $f$ and $g$ differ by a polynomial. In this special case, transfer learning can be highly beneficial.
We refer to $g$ as a ``perfect reference" of $f$ when the bias strength is zero, and an ``imperfect reference" otherwise.

There are two natural and important goals for  transfer learning: to accurately quantify the contribution of observations from the source domain to the regression task in the target domain, and to develop an optimal transfer learning algorithm. The answer to the first objective depends on various factors, including sample sizes $n_P$ and $n_Q$, the smoothness of $f$ and $g$, and the bias strength $\epsilon$. In this paper, we investigate the \emph{non-asymptotic} minimax risk of this problem and propose a data-driven, adaptive transfer learning algorithm to achieve optimal results.

\subsection{Main Results and Our Contribution}

We first establish the minimax optimal rate of convergence for transfer learning for nonparametric regression in the  posterior drift setting. Suppose $f$ is $\beta_Q$-smooth and $g$ is $\beta_P$-smooth (which will be defined precisely later). Let $F$ denote the set of distribution pairs $(Q, P)$ defined in (\ref{eqn: F set}) in Section \ref{section: problem formulation}. It is shown that the minimax risk of transfer learning satisfies 
 \begin{align*}
&C_L\cdot\bigg(n_{\max}^{-\frac{2\bmax}{2\bmax+d}}+ (\epsilon \wedge n_Q^{-\frac{\beta_Q}{2\beta_Q+d}})\cdot n_Q^{-\frac{\beta_Q}{2\beta_Q+d}}+\frac{1}{n_Q}\bigg)\leq \inf_{\hat f}\sup_{(Q, P)\in F}\E ||\hat f-f||_2^2\\
&\leq C_U\cdot\bigg( n_{\max}^{-\frac{2\bmax}{2\bmax+d}}\ln^4(n_{\max})+ \\
&\ln^8(n_Q)\cdot(\epsilon \wedge n_Q^{-\frac{\beta_Q}{2\beta_Q+d}})\cdot n_Q^{-\frac{\beta_Q}{2\beta_Q+d}}+\frac{\ln^4(n_Q)}{n_Q}\bigg),\end{align*}
for some positive constants $C_L$ and $C_U$ not depending on $\epsilon$, $n_P$ or $n_Q$, where $n_{\max}=\max( n_{P},n_Q)$ and $\beta_{\max}=\max(\beta_Q,\beta_P)$. In the special case of observing the data from the target domain only, i.e., $n_P=0$,  then the minimax risk for estimating $f$ is of order $n_Q^{-\frac{2\beta_Q}{2\beta_Q+d}}$. Comparing it with the transfer learning risk, we can conclude when and how much transfer learning is helpful for the target task. The necessary  and sufficient condition for transfer learning to improve the estimation performance is  that the bias strength is smaller than $n_Q^{-\frac{\beta_Q}{2\beta_Q+d}}$ and either the mean function from the source domain is smoother or the sample size of the source domain is larger. If we fix $f$ and $g$ and thus the bias strength $\epsilon$ and let $n_Q$ go to infinity then this condition fails when $n_Q$ is sufficiently large unless $\epsilon=0$. This means in order to make transfer learning work \emph{asymptotically}, $g$ has to be a perfect reference. However, with any fixed and finite sample size, the non-asymptotic analysis above shows that transfer learning can help with an imperfect reference function $g$ as long as the bias strength $\epsilon$ is smaller than the phase transition threshold $n_Q^{-\frac{\beta_Q}{2\beta_Q+d}}$. 

There are some interesting phenomena from the minimax analysis in the case where  bias strength $\epsilon$ is sufficiently small. 
\begin{enumerate}
    \item If the function $g$ is rougher than $f$, i.e., $\beta_P < \beta_Q$,  then the minimax risk does not depend on $\beta_P$ and is the same as the minimax risk when $\beta_P=\beta_Q$. This means that transfer learning can be effective even if $g$ is less smooth than $f$.
    \item If estimation of $f$ and $g$ is considered separately using the data from the target domain and from the source domain alone, the usual minimax risks for estimating $f$ and $g$ are proportional to $n_Q^{-\frac{2\beta_Q}{2\beta_Q+d}}$ and  $n_P^{-\frac{2\beta_P}{2\beta_P+d}}$ respectively. Then if $\beta_P < \beta_Q$ and $n_P\gg n_Q$ or $\beta_P > \beta_Q$ and $n_P\ll n_Q$, the minimax risk for the transfer learning can be much smaller than either of these two minimax risks, provided $\epsilon$ is sufficiently small. This phenomenon sheds new light on the understanding of transfer learning in that even the task from the source domain is harder, i.e., $n_Q^{-\frac{2\beta_Q}{2\beta_Q+d}}\ll n_P^{-\frac{2\beta_P}{2\beta_P+d}}$, it may still help the task in the target domain. 
\end{enumerate}
A novel transfer learning algorithm is developed and shown to attain the minimax optimal risk, possibly up to a logarithmic factor. 
However, the algorithm relies on knowledge of the smoothness parameters $\beta_Q$ and $\beta_P$. To address this, we propose a data-driven algorithm that adaptively achieves the minimax risk, up to a logarithmic factor, over a wide range of parameter spaces. Simulation studies are conducted to further demonstrate the performance of the adaptive transfer learning algorithm and validate the phenomena discussed.

Simulation studies are conducted to evaluate the performance of the adaptive transfer learning algorithm. The numerical results  further support our theoretical analysis. The proposed method is then applied to a wine quality dataset \citep{cortez2009modeling} to compare the performance of direct local polynomial regression on red wine data to using the transfer learning algorithm on both red and white wine data with varying numbers of observations. The results show that the transfer learning method improves performance.

The results and algorithms can also be extended to the setting of multiple source distributions. Suppose there are $K$ source distributions ($P_1, \dots, P_K$) and one target distribution $Q$. Each source distribution $P_j$ corresponds to a mean function $g_j$ and the difference between $f$ and each $g_j$ can be well approximated by a polynomial function of a given order. We establish the non-asymptotic minimax risk and construct an adaptive procedure that simultaneously attains the optimal risk, up to a logarithmic factor, over a large collection of parameter spaces.

\subsection{Related Literature}

The problem of transfer learning for nonparametric regression in the posterior drift setting has been studied in \citet{wang2016nonparametric}, under the assumption that the mean functions from both domains have the same Sobolev smoothness and the difference belongs to a smoother Sobolev class. An upper bound for the performance of transfer learning was obtained, however, no lower bound was provided, leaving it unclear if the upper bound is sharp. Additionally, their upper bound can only be achieved when the smoothness is known, which is not typically the case in practice and no adaptation to smoothness was considered. In contrast, in the present paper, we allow the mean functions to have different \holder smoothness and assume that the difference function can be approximated in $L_1$ distance by a polynomial function. We prove matching upper and lower bounds, up to a logarithmic factor, to quantify what transfer learning can achieve. In the covariate shift setting, transfer learning for nonparametric regression has also been considered in \cite{huang2006correcting, wen2014robust}.

For nonparametric transfer learning, much attention has been given to classification, with the general problem being studied in \citet{ben2007analysis}, \citet{blitzer2008learning}, \citet{mansour2009domain}. Theoretical results and adaptive procedures have been developed in both the posterior drift setting \citep{cai2021transfer, reeve2021adaptive} and the covariate shift setting \citep{shimodaira2000improving, sugiyama2007direct}.

The transfer learning problem we consider here is related to the classical nonparametric regression literature, where only observations from the target domain are available. In particular, our algorithm uses local polynomial regression as a basic tool, which has been well-studied in the conventional setting (see, for example, \citet{stone1977consistent}, \citet{cleveland1979robust}, \citet{tsybakov1986robust}, \citet{fan1992variable}, \citet{fan1993local}, and \citet{xiao2003more}).

\subsection{Organization and Notation}

The rest of the paper is organized as follows. We finish this section with the notation. Section \ref{section: problem formulation} presents the precise formulation of the transfer learning problem studied in the paper. We then construct a novel algorithm in Section \ref{section: minimax risk} and derive an upper bound for its risk. A matching lower bound is then established.   The problem of adaptation to smoothness is considered  in Section \ref{section: adaptive alg}. A data-driven procedure is proposed and shown to be adaptively rate-optimal. Simulation studies and a real application are carried out in Section \ref{section: simulation} and \ref{section: application} to investigate the numerical performance of the adaptive algorithm. Section \ref{section: multiple} generalizes the methods and theoretical analysis to the multiple source distributions setting and Section \ref{section: discussion} discusses possible future directions.  For reasons of space, the proofs  are given in the Supplementary Material \citep{CaiPu}.

The following notation will be used in the paper. For any function $h:[0,1]^d \rightarrow \R$, let $||h||_1=\int_{[0,1]^d} |h(x)|dx $, $||h||=||h||_2=\sqrt{\int_{[0,1]^d}h^2(x)dx}$ and $||h||_\infty=\sup_{x\in [0,1]^d}|h(x)|$.
For any $a, b \in \R$, we define $a\vee b= \max(a,b)$ and $a\wedge b = \min(a,b)$. For any $\beta>0$, let $w(\beta)$ denote the largest integer that is strictly smaller than $\beta$. Let $\text{Poly}(M, T)$ denote the polynomial functions whose degree is smaller than or equal to $T$ and  coefficients have absolute values upper bounded by $M$. For any multi-index $\t=(t_1,\dots,t_d)$ and  vector $x=(x_1,\dots,x_d)$, let $x^{\t}=\prod_{i=1}^d x_i^{t_i}$. For any multi-index $\t=(t_1,\dots,t_d)$, define $|\t|=\sum_{j=1}^d t_j$. Define the multi-index class $\Lambda(k)=\{\t:|\t|\leq k\}$.
For two functions $h_1(n), h_2(n)>0$, we write $h_1(n)=\O\big(h_2(n)\big)$ if $\limsup_{n\rightarrow \infty}\frac{h_1(n)}{h_2(n)}<\infty$;  $h_1(n)=\tilde{\O}\big(h_2(n)\big)$ if there exists a constant $C>0$ such that $h_1(n)=\O\big(h_2(n)\cdot \ln^{C}(n)\big)$; 
$h_1(n)=\Omega\big(h_2(n)\big)$ if  $\liminf_{n\rightarrow \infty}\frac{h_1(n)}{h_2(n)}>0$; $h_1(n)=\Theta\big(h_2(n)\big)$ if $h_1(n)=\O\big(h_2(n)\big)$ and $h_2(n)=\O\big(h_1(n)\big)$; and $h_1(n)=\tilde{\Theta}\big(h_2(n)\big)$ if $h_1(n)=\tilde{\O}\big(h_2(n)\big)$ and $h_2(n)=\tilde{\O}\big(h_1(n)\big)$. 

\section{Problem Formulation}
\label{section: problem formulation}

We begin by formally defining the \holder smoothness as follows.
\begin{definition}(\holder class)
For any positive numbers $\beta,L$, the \holder function class $\H(\beta,L)$  is defined to be the set of all  functions $h$ with continuous partial derivatives of order  $w(\beta)$ that satisfy
$$\max_{|\t|\leq w(\beta)}\sup_{x\in [0,1]^d}|D^{\t}h(x)|+\max_{|\t|=w(\beta)}\sup_{x\neq y\in [0,1]^d}\frac{|D^{\t}h(x)-D^{\t}h(y)|}{||x-y||^{\beta-|\t|}}\leq L.$$
\end{definition}

We model the target mean function $f$ and the source mean function $g$ as \holder smooth functions and assume for some  constants $\beta_Q,\beta_P,L_Q,L_P>0$, 
$$f\in \H(\beta_Q,L_Q) \quad {\rm and} \quad g\in \H(\beta_P,L_P).$$
In this case, we call $f$ $\beta_Q$-smooth and $g$ $\beta_P$-smooth.
 
\begin{definition}
For any $\epsilon>0,M>0$, positive integer  $T$, the class of functions $\Psi(\epsilon,M,T)$ is defined as all the functions $h$  such that  there exists a polynomial function $\psi(x)\in \text{Poly}(M, T)$  such that 
$$||\psi-h||_1= \int_{[0,1]^d} |\psi(x)-h(x)| dx \leq \epsilon.$$
\end{definition}
We assume $f-g\in \Psi(\epsilon,M,T)$. This assumption requires that $g$ is close in $L_1$ distance to $f$ plus a polynomial of order $T$. In this paper we only consider  polynomials with coefficients  bounded in absolute value by a constant $M$. It is possible to generalize $\psi$ to be an arbitrary polynomial function. The discussion on this generalization is deferred to Section \ref{section: discussion}.

\begin{definition}
For any $u_1,u_2>0$, the class of sub-Gaussian random variables $G(u_1,u_2)$ with constants $u_1,u_2>0$ are defined as all random variables $Z$ such that for any $t>0$,
$$\P\Big(|Z|\geq t\Big)\leq u_1 \cdot e^{-u_2 t^2}.$$
\end{definition}
For any $x\in [0,1]^d$ we assume the random noises $z_1|X_1=x$ and $z_1'|X_1'=x$ to be sub-Gaussian with some constants $u_1,u_2$. This assumption  ensures the outcome to be not heavy-tailed. We assume the marginal distributions of $X_1$ and $X_1'$ have density functions $f_{den}^Q$ and $f_{den}^P$ respectively and they are  lower and upper bounded by constants $C^L,C^U$.

Given these definitions, the parameter space is defined by 
\begin{align}
&F(\beta_Q,\beta_P,\epsilon,u_1,u_2,M,T,{L_P},L_Q,C^L,C^U)=\bigg\{(Q,P):f\in \H(\beta_Q,L_Q),\notag\\
&g\in \H(\beta_{P},L_{P}), f-g\in\Psi(\epsilon,M,T),z_1 |X_1,z_{1}' |X_1'\in G(u_1,u_2),C^{L}\leq\\
&f_{den}^Q,f_{den}^P\leq C^U\bigg\}\label{eqn: F set}.
\end{align}
This space will be denoted by $F(\beta_Q,\beta_P,\epsilon)$ when there is no confusion. 
The minimax estimation risk over this parameter space is then defined as
\begin{align*}
&R_{\beta_Q,\beta_P,\epsilon,u_1,u_2,M,T,L_P,L_Q,C^L,C^U}(n_Q,n_{P})=\\
&\inf_{\hat f} \sup_{(Q,P)\in F(\beta_Q,\beta_P,\epsilon)} \E \ (\hat f(X)-f(X))^2. 
\end{align*}

For simplicity we may write  $R_{\beta_Q,\beta_P,\epsilon,u_1,u_2,M,T,L_P,L_Q,C^L,C^U}(n_Q,n_{P})$ as $R_{\beta_Q,\beta_P,\epsilon}$ or $R(n_Q,n_P)$ if there is no confusion. It is interesting to understand when and how much transfer learning can improve the estimation accuracy in the target domain. This question can be answered by comparing the transfer learning minimax risk $R_{\beta_Q,\beta_P,\epsilon}$ with the 
minimax risk using only data from the target domain, which is well known to be of order $n_Q^{-\frac{2\beta_Q}{2\beta_Q+d}}$.

\section{Minimax Risk}
\label{section: minimax risk}

We consider in this section the minimax risk for the transfer learning problem.  We begin with a brief review of local polynomial regression  in Section \ref{section: lpr} that serves as a basic tool for nonparametric regression in our algorithm. We then formally present in Section \ref{section: alg} the algorithm. The minimax risk is established in Section \ref{section: small minimax rate}  and a discussion on interesting phenomena is given in Section \ref{section: discussion of minimax}.

\subsection{Local Polynomial Regression}
\label{section: lpr}

Local polynomial regression has been widely recognized for its empirical success and desirable theoretical properties  (\citealp{cleveland1988locally,fan1992variable,fan1993local}).
In particular, local polynomial regression achieves the minimax optimal rate over a \holder ball with properly tuned parameters (\citealp{gyorfi2002distribution}). 

For observations $\D=\{(X^{(i)},Y^{(i)}),i=1,\dots,n\}$, degree $l$ and bandwidth $b$ (we assume $1/b$ is an integer) the local polynomial regression estimate is defined as follows. Divide $[0,1)^d$ into $\frac{1}{b^d}$ hypercubes $\{\prod_{i=1}^d [b\cdot a_i,b\cdot a_i+b):a_i=0,\dots,1/b-1\}$. For each hypercube $B(a_1,\dots,a_d)=\prod_{i=1}
^d [b\cdot a_i,b\cdot a_i+b)$, let all the observations whose covariates falling into this hypercube be $\{(X_{i_1},Y_{i_1}),\dots,(X_{i_r},Y_{i_r})\}$. Let $a_{\rm mid}=(a_1+b/2,\dots,a_d+b/2)$ be the center of this hypercube. The local polynomial regression estimate $\hat f_{\rm lpr}$ on $B(a_1,\dots,a_d)$ is given by
$$\lpr(x;\D,l,b)=\sum_{\t\in \Lambda(l)}c_{\t}^{\rm lpr}(\frac{x-a_{\rm mid}}{b/2})^{\t}, \forall x\in B(a_1,\dots,a_d),$$
where $\{c_{\t}^{\rm lpr}:\t\in \Lambda(l)\}$ are given by $$\{c_{\t}^{\rm lpr}:\t\in \Lambda(l)\}=\argmin_{\{c_\t:\t\in \Lambda(l)\}}\sum_{m=1}^r (Y_{i_m}-\sum_{\t\in \Lambda(l)}c_{\t}(\frac{X_{i_m}-a_{\rm mid}}{b/2})^{\t})^2.$$

The confidence upper and lower limits are constructed as follows:
$$\lpr^{ub}(x;\D,l,b,\beta)=\lpr(x:\D,l,b)+\sqrt{\ln(|\D|)}\cdot (b^{-\beta}+\frac{\ln^2(|\D|)}{\sqrt{|\D|\cdot b^d}}),$$
$$\lpr^{lb}(x;\D,l,b,\beta)=\lpr(x:\D,l,b)-\sqrt{\ln(|\D|)}\cdot (b^{-\beta}+\frac{\ln^2(|\D|)}{\sqrt{|\D|\cdot b^d}}).$$
The length of confidence interval is then
$$L_{\rm CI,lpr}(\D,l,b,\beta)=2\cdot \sqrt{\ln(|\D|)}\cdot (b^{-\beta}+\frac{\ln^2(|\D|)}{\sqrt{|\D|\cdot b^d}}).$$

\subsection{The Confidence Thresholding(CT) Algorithm}
\label{section: alg}

We now present our main transfer learning algorithm. Since we have data from both the source and target domains, the most important and challenging step of this algorithm is integrating the information from both domains. This step in our algorithm is called confidence thresholding. We shall first present this confidence thresholding procedure and then show the details of our algorithm.

\subsubsection{The Confidence Thresholding Estimator}
\label{section: ct}

\begin{figure}[htbp]
    \centering
\begin{minipage}[t]{0.48\textwidth}
{\includegraphics[scale=0.35]{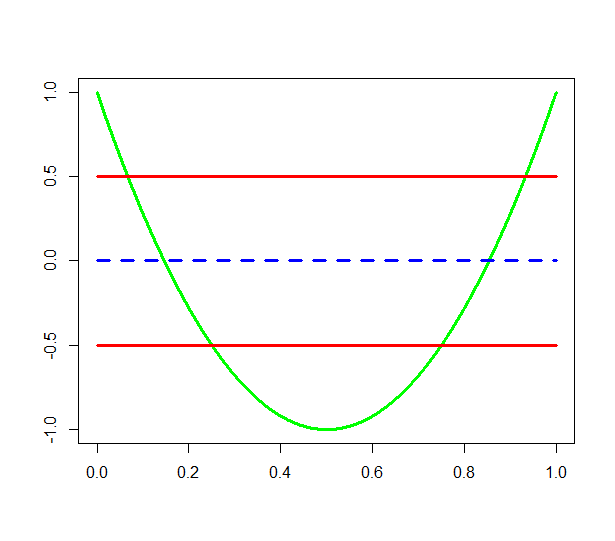}}
\end{minipage}
\begin{minipage}[t]{0.48\textwidth}
\includegraphics[scale=0.35]{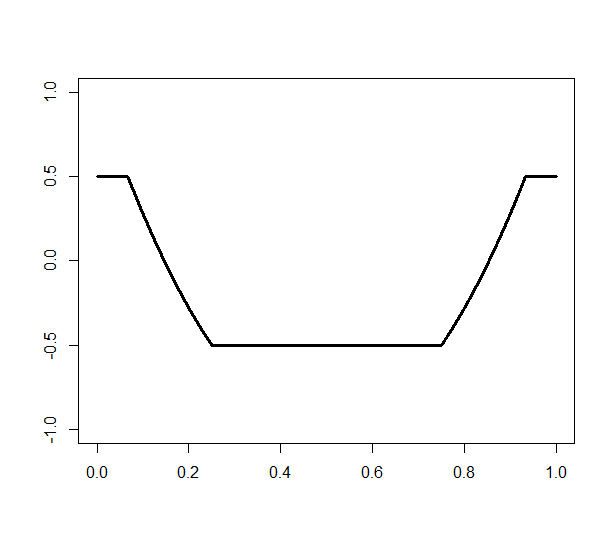}
\end{minipage}
\caption{An illustration of the confidence thresholding estimator. On the left panel, the blue dashed line is $\hat h_1$, the green line is $\hat h_2$ and two red lines are the confidence upper bound $\hat h_1+e_1$ and lower bound $\hat h_1-e_1$. On the right panel, the black line is the confidence thresholding estimator $ \hat \mu_{\rm ct}(\hat h_1(x),\hat h_2(x), e_1)$. }
\label{fig: confidence thresholding}
\end{figure}

We first introduce the confidence thresholding estimator that is designed to estimate a function when one has access to two different estimates. Suppose we have two different estimators $\hat h_1$ and $\hat h_2$ for some unknown function $h$. $\hat h_1$ converges slower and $\hat h_2$ converges faster but is slightly biased, which means $\hat h_2$ converges to a function that is different from but close to $h$ in $L_1$ distance. The confidence thresholding estimator is constructed based on $\hat h_1$ and $\hat h_2$ as follows. Let $e_1$ be an upper bound of the $L_\infty$ norm of $\hat h_1-h$. Then a ``confidence interval" for $h(x)$ is $[\hat h_1(x)-e_1,\hat h_1(x)+e_1]$ for all $x\in [0,1)^d$. There are three different possible cases of the relationship between this confidence interval and $\hat h_2(x)$. If $\hat h_2(x)$ is greater than the upper bound of the confidence interval, then this confidence interval upper bound is better than $\hat h_2(x)$ in estimating $h(x)$ and in this case the confidence interval upper bound is used as the estimate. If  $\hat h_2(x)$ is in the confidence interval, then $\hat h_2(x)$ is acceptable and we use $\hat h_2(x)$. If $\hat h_2(x)$ is smaller than the lower bound of the confidence interval, then the confidence interval lower bound is better than $\hat h_2(x)$ in estimating $h(x)$ and the  confidence interval lower bound is used as the estimate. We call this estimator the \emph{confidence thresholding estimator}:
\begin{align}\label{eqn: confidence thr}
    \hat \mu_{\rm ct}(\hat h_1(x),\hat h_2(x), e_1)=  \hat h_1(x) + {\rm sgn}(\hat h_2(x)-\hat h_1(x))\cdot (|\hat h_2(x)-\hat h_1(x)|\wedge e_1).
\end{align}
See Figure \ref{fig: confidence thresholding} for an illustration of the confidence thresholding estimator.
 
The following lemma justifies the convergence of $\hat \mu_{\rm ct}$.
\begin{lemma}\label{lemma: st}
Suppose for a function $h:[0,1]^d\rightarrow \R$, we have two estimates $\hat h_1$ and $\hat h_2$. Suppose for some $e_1,e_2,e_2'>0$, $||h-\hat h_1||_\infty \leq e_1$ and
$||h+\tilde h-\hat h_2||_\infty \leq e_2\leq e_1$ where function $\tilde h:[0,1]^d\rightarrow \R$ satisfies $||\tilde h||_1\leq e_2'$. Let $\hat h(x)= \hat \mu_{\rm ct}(\hat h_1(x),\hat h_2(x),e_1)$ for all $x\in [0,1]^d$. Then $$||\hat h-h||_2\leq (e_2+\sqrt{4e_1\cdot e_2'})\wedge 2e_1.$$
\end{lemma}
With this lemma we can compare the confidence thresholding estimator with the two original estimators. In the setting of this lemma, the $L_2$ error of $\hat h_1$ is upper bounded by $e_1$ and the $L_2$ error of $\hat h_2$ can be arbitrarily large since $||\tilde h_2||_2$ can be arbitrarily large. The $L_2$ error of the confidence thresholding estimator is upper bounded by $2e_1$, so it is at least as good as the $\hat h_1$ up to a constant. Besides, in the case where $e_2'\ll e_1$ and $e_2\ll  e_1$, the confidence thresholding estimator outperforms both of two original estimators. 

\subsubsection{The Confidence Thresholding(CT) Algorithm} 

We now present in detail the CT algorithm, which utilizes the confidence thresholding estimator and involves fitting local polynomial regression twice to produce two preliminary estimators. These estimators are used to mimic $\hat h_1$ and $\hat h_2$ in the confidence thresholding estimator.

We begin with sample splitting by randomly dividing $\D_Q$  into two equal-sized subsets $\D_{Q,1}$ and $\D_{Q,2}$. We first fit local polynomial regression on $\D_{Q,1}$ with some bandwidth $\tilde \beta_Q$ and obtain an estimate $\hat f_{\rm ref}$. We also construct a confidence interval and compute its length, which is denoted by $L_{CI}$. In the confidence thresholding estimator, $\hat f_{\rm ref}$ will serve as $\hat h_1$ and $L_{CI}/2$ will serve as $e_1$.
We then fit another local polynomial regression to mimic $\hat h_2$. We can fit it either on $\D_{Q,1}$ or $\D_{P}$ because $\hat h_2$ is allowed to be biased. To get a faster convergence, we fit this local polynomial regression  on the dataset  with some larger bandwidth $\tilde b$. Let this estimate be $\hat f_{\rm raw}$.

Note $f-g$ is close to a polynomial function in $L_1$ distance. Then $\hat f_{\rm raw}$ plus some polynomial function $\psi$ should be close to $f$. If $\psi$ were known,  the confidence thresholding estimator 
$\hat f_{ct}(\cdot;\psi)=\hat \mu_{\rm ct}(\hat f_{\rm ref},\hat f_{\rm raw}+\psi,L_{CI}/2)$ could be used to estimate $f(x)$. However, since $\psi$ is unknown, we shall first estimate it  and then plug the estimate into the confidence thresholding estimator. The estimator $\hat \psi$ is obtained by minimizing the empirical mean squared error on the validation set $\D_{Q,2}$. Formally,
$$\hat \psi=\argmin_{\psi\in \text{Poly}(\sqrt{\ln(n_Q)}, l)}\sum_{i= \lfloor\frac{n_{Q}}{2}\rfloor+1}^{n_Q}[Y_{i}-\hat f_{\rm ct}(X_i;\psi)]^2.$$
Finally, we truncate the estimate  $\hat \mu_{\rm ct}(\hat f_{\rm ref}(x),\hat f_{\rm raw}+\hat \psi,L_{CI}/2)$ since  $f$ is bounded.  The CT algorithmx is summarized in Algorithm \ref{alg: non}.

\begin{algorithm} 
\caption{$\A^{\ref{alg: non}}(\tilde \beta_Q,\tilde \beta_P, l)$} \label{alg: non}
\vspace*{0.12 cm}
\KwIn{ \holder smoothness $\tilde \beta_Q$, $\tilde \beta_P$ and polynomial degree $l$.}
\vspace*{0.24 cm}
Split $\D_Q$ into $\D_{Q,1}=\{(X_1,Y_1),\dots,(X_{\lfloor \frac{n_Q}{2}\rfloor},Y_{\lfloor \frac{n_Q}{2}\rfloor})\}$ and 
$\D_{Q,2}=\{(X_{\lfloor \frac{n_Q}{2}\rfloor+1},Y_{\lfloor \frac{n_Q}{2}\rfloor+1}),\dots,(X_{n_Q},Y_{n_Q})\}$.
\\
\vspace*{0.12 cm}

Calculate $\tilde \beta_{\max}=\max(\tilde \beta_Q,\tilde \beta_P)$, $n_{Q,1}=|\D_{Q,1}|$, $n_P=|\D_{P}|$, $\tilde n_{\max}=\max(n_{Q,1}, n_P)$, $\tilde b=\frac{1}{\lfloor \tilde n_{\max} ^{\frac{1}{2\tilde \beta_{\max}+d}}\rfloor}$, $\tilde b_Q=\frac{1}{\lfloor n_{Q,1} ^{\frac{1}{2\tilde \beta_Q+d}}\rfloor}$. \\
\vspace*{0.12 cm}
Fit local polynomial regression on $\D_{Q,1}$ with bandwidth $\tilde b_{Q}$, and calculate estimates and confidence interval length.\\
$$\hat f_{\rm ref}(\cdot)=\lpr(\cdot;\D_{Q,1},l,\tilde b_Q),$$ $$L_{CI}=L_{\rm CI,lpr}(\D_{Q,1},l,\tilde b_Q,\tilde \beta_Q).$$
\vspace*{0.12 cm}
\If{$n_{Q,1}>n_P$}{
fit local polynomial regression on $\D_{Q,1}$ with bandwidth $\tilde b$, $\hat f_{\rm raw}(\cdot)=\lpr(\cdot;\D_{Q,1},l, \tilde b)$,\\}
\Else{fit local polynomial regression on $\D_{P}$ with bandwidth $\tilde b$, $\hat f_{\rm raw}(\cdot)=\lpr(\cdot;\D_{P},l, \tilde b)$.\\}

\vspace*{0.12 cm}
Estimate $\hat \psi$ by 
$$\hat \psi=\argmin_{\psi\in \text{Poly}(\sqrt{\ln(n_Q)}, l)}\sum_{i=n_{Q,1}+1}^{n_Q}[Y_{i}-\hat f_{\rm ct}(X_i;\psi)]^2,$$
where $\hat f_{\rm ct}(x;\psi)=\tilde \mu_{\rm ct}(\hat f_{\rm ref}(x),\hat f_{\rm raw}(x)+\psi(x),L_{CI}/2)$.\\
\vspace*{0.12 cm}
Truncate the estimate at $n_Q$,
$$\hat f(x)={\rm sgn}(\hat f_{\rm ct}(x;\hat \psi))\cdot (|\hat f_{\rm ct}(x;\hat \psi)|\wedge n_Q)$$
\end{algorithm}

\begin{remark}{\rm 
The bandwidths $\tilde b_Q$ and $\tilde b$ are chosen such that $\tilde b_Q\propto n_Q ^{-\frac{1}{2\tilde\beta_Q+d}}$ and $ \tilde b\propto(n_Q\vee n_P) ^{-\frac{1}{2\tilde \beta_{\max}+d}}$, due to the fact that $n^{-\frac{1}{2\beta+d}}$ is the optimal bandwidth for estimating a $\beta$-smooth function based on $n$ observations   \citep{gyorfi2002distribution}. 
}\end{remark}

\subsection{Minimax Risk}
\label{section: small minimax rate}

The following theorem gives an upper bound for the risk of the CT algorithm. Recall $\beta_{\max}=\beta_Q \vee \beta_P$ and $n_{\max}=n_P \vee n_Q$.
\begin{theorem}[Minimax upper bound]\label{th: non adaptive}
Suppose in Algorithm \ref{alg: non}, $\tilde \beta_Q=\beta_Q,\tilde \beta_P=\beta_P,l\geq w(\beta_{\max})\vee T$, then the risk of this algorithm satisfies
\begin{align*}
&R(\hat f)= \sup_{(Q,P)\in F(\beta_Q,\beta_P,\epsilon)}  \E (\hat f(X)-f(X))^2\leq \\
&C_U\cdot\bigg( n_{\max}^{-\frac{2\bmax}{2\bmax+d}}\ln^4(n_{\max})+ \ln^8(n_Q)\cdot(\epsilon \wedge n_Q^{-\frac{\beta_Q}{2\beta_Q+d}})\cdot n_Q^{-\frac{\beta_Q}{2\beta_Q+d}}+\frac{\ln^4(n_Q)}{n_Q}\bigg),    
\end{align*}

for some constant $C_U>0$ that only depends on $\beta_Q,\beta_P$,$u_1,u_2$,$M,T,d,l$,$L_P,L_Q,C^L,C^U$ and not on $n_Q,n_{P},\epsilon$.
\end{theorem}
The next theorem provides a lower bound for the minimax risk and shows that CT algorithm is minimax optimal up to a logarithmic factor.

\begin{theorem}[Minimax lower bound]\label{th: lower bound}  There exists a constant $C_L>0$ that only depends on $\beta_Q,\beta_P,u_1,u_2,M,T,d,L_P,L_Q$ and not on $n_Q,n_{P},\epsilon$ such that
 $$R_{\beta_Q,\beta_P,\epsilon} \geq C_L\cdot\bigg( n_{\max}^{-\frac{2\bmax}{2\bmax+d}}+ (\epsilon \wedge n_Q^{-\frac{\beta_Q}{2\beta_Q+d}})\cdot n_Q^{-\frac{\beta_Q}{2\beta_Q+d}}+\frac{1}{n_Q}\bigg).$$
\end{theorem}

Theorems \ref{th: non adaptive} and \ref{th: lower bound} together show that the non-asymptotic minimax risk of  transfer learning for nonparametric regression $R_{\beta_Q,\beta_P,\epsilon}$ is proportional to
\begin{align}
n_{\max}^{-\frac{2\bmax}{2\bmax+d}}+ (\epsilon \wedge n_Q^{-\frac{\beta_Q}{2\beta_Q+d}})\cdot n_Q^{-\frac{\beta_Q}{2\beta_Q+d}}+\frac{1}{n_Q}.\label{eqn: minimax}
\end{align}
Comparing this risk with the minimax risk of nonparametric regression with the observations from the target domain only, which is proportional to $n_Q^{-\frac{2\beta_Q}{2\beta_Q+d}}$, we can see when and how transfer learning improve the estimation accuracy for $f$. The sufficient and necessary condition is that the bias strength $\epsilon\ll n_Q^{-\frac{\beta_Q}{2\beta_Q+d}}$  and either the source domain has  a smoother mean function $\beta_P>\beta_Q$ or much more observations $n_P\gg n_Q$. 

The second term $(\epsilon \wedge n_Q^{-\frac{\beta_Q}{2\beta_Q+d}})\cdot n_Q^{-\frac{\beta_Q}{2\beta_Q+d}}$ in (\ref{eqn: minimax})
represents the influence of the bias strength to the difficulty of the current problem. It has two phase transition points. The first is $n_Q^{-\frac{\beta_Q}{2\beta_Q+d}}$.  If the bias strength is larger than it then the minimax risk (\ref{eqn: minimax}) is as large as the minimax risk of regression with the target domain data only, which is proportional to $n_Q^{-\frac{\beta_Q}{2\beta_Q+d}}$. If the bias strength is smaller than it then whether the minimax risk (\ref{eqn: minimax}) is smaller than $n_Q^{-\frac{\beta_Q}{2\beta_Q+d}}$ does not depend on the bias strength. In other words, $n_Q^{-\frac{\beta_Q}{2\beta_Q+d}}$ is the maximum tolerable bias strength for transfer learning to help and quantifies the robustness of this model. The second phase transition point is $n_Q^{-\frac{\beta_Q+d}{2\beta_Q+d}}\vee n_Q^{\frac{\beta_Q}{2\beta_Q+d}}\cdot n_{\max}^{-\frac{2\beta_{\max}}{2\beta_{\max}+d}}$. Whether the bias strength is larger than it determines whether the influence of the bias strength is the dominating term. In other words, if the bias strength is smaller than it then transfer learning can work as if there is no bias.

The first term in equation (\ref{eqn: minimax}) is equivalent to the minimax rate for nonparametric regression over a $\beta_{\max}$-smooth \holder class with $n_{\max}$ observations. This suggests that transfer learning can benefit from larger sample sizes and improved smoothness, regardless of whether these advantages are present in different domains. Essentially, transfer learning allows for the transfer of sample size and smoothness to a common domain.
\subsection{Discussion}
\label{section: discussion of minimax}

We now take a closer look at the minimax risk in cases where the bias strength is strong enough to not be the dominant term in the minimax risk.We explore two unique phenomena displayed by the minimax risk: {\it auto-smoothing} and {\it super-acceleration}.

\begin{itemize} 

 \item \textbf{Auto-smoothing:}  When $\beta_P<\beta_Q$, the minimax rate  is
  $$ n_{\max}^{-\frac{2\beta_Q}{2\beta_Q+d}}+ (\epsilon \wedge n_Q^{-\frac{\beta_Q}{2\beta_Q+d}})\cdot n_Q^{-\frac{\beta_Q}{2\beta_Q+d}}+\frac{1}{n_Q},$$ which does not depend on $\beta_P$. This implies that even if $g$ is highly irregular ($\beta_P\approx 0$), it is still possible to estimate $f$ as if $g$ is a $\beta_Q$-smooth function. The CT algorithm only relies on $\tilde \beta_Q$ and $\tilde \beta_{\max}$, thus it is not affected by $\tilde \beta_P$ if $\tilde \beta_P \leq \tilde \beta_Q$. This aligns with the auto-smoothing phenomenon observed in minimax theory.
 
\item \textbf{Super-acceleration:} In transfer learning, a common question of interest is whether and to what extent observations from the source domain can significantly improve estimation accuracy in the target domain. In this scenario, if the source domain has a smoother mean function but a smaller sample size, i.e. $\beta_P > \beta_Q$ and $n_P < n_Q$, and the bias strength $\epsilon$ is sufficient, then the minimax risk for transfer learning is $R_{\beta_Q,\beta_P,\epsilon}= n_Q^{-\frac{2\beta_P}{2\beta_P+d}}$, which is smaller than both the minimax risk for estimating $f$ using data from the target domain alone and the minimax risk for estimating $g$ using data from the source domain alone. This phenomenon is referred to as  {\it super-acceleration}.
This provides new insights into transfer learning by demonstrating that it can significantly enhance performance on the target domain even if the task in the source domain is more difficult (based on data from the source domain alone). Similarly, super-acceleration also occurs if the source domain has a rougher mean function and more observations.

On the other hand, if the source domain has a smoother mean function and a larger sample size, it is not surprising that transfer learning can improve the convergence rate. In this case, $f$ can be estimated as accurately as $g$, as the minimax risk for transfer learning
is of order $n_P^{-\frac{2\beta_P}{2\beta_P+d}}$ when $\epsilon$ is sufficiently small. There have been other results in transfer learning for different tasks where the best one can do on the target domain is as good as the performance of the corresponding task on the source domain. This kind of acceleration is referred to as {\it normal acceleration}.
The following table summarizes different cases.
\end{itemize}

\vspace{-15pt}
\begin{center}
\begin{tabular}{|c|c|c|c|}
\hline
  & $\beta_Q > \beta_P$ & $\beta_Q=\beta_P$ & $\beta_Q<\beta_P$ \\
\hline
$n_Q\gg n_{P}$  &no acceleration & no acceleration & super-acceleration\\
\hline 
$n_Q\propto n_{P}$ &no acceleration &no acceleration & normal acceleration\\
\hline
$n_Q\ll  n_{P}$ & super-acceleration& normal acceleration &normal acceleration\\
\hline
\end{tabular}
\end{center}

\section{Adaptive Confidence Thresholding Algorithm}
\label{section: adaptive alg}

Section \ref{section: minimax risk} establishes  the non-asymptotic minimax risk  for estimation over the parameter space $F(\beta_Q,\beta_P,\epsilon)$ and the optimality of the CT algorithm. However, the CT algorithm requires the knowledge of the smoothness parameters $\beta_Q$ and $\beta_P$, which are typically unknown. A natural and  important question is whether it is possible to construct a data-driven algorithm that adaptively achieves the optimal risk simultaneously over a wide rage of parameter spaces.

In this section we develop an adaptive algorithm, called adaptive confidence thresholding (ACT) algorithm, that is based on the CT algorithm and consists of three main steps:

\begin{itemize}
\item \textbf{Step 1: Constructing a set of smoothness parameter pairs.} Since the CT algorithm only depends on $\tilde \beta_Q$ and $\tilde \beta_{\max}$, we construct a finite set $S_Q\subset \R$ for  $\tilde \beta_Q$'s and a finite set $S_{\max}$ for $\tilde \beta_{\max}$'s. $\tilde \beta_Q$ and $\tilde \beta_{\max}$ are to be chosen from $S_Q$ and $S_{\max}$ respectively. $S_Q$ and $S_{\max}$ are both arithmetic sequences. The common differences of these two sequences are roughly proportion to $\frac{1}{\ln(n_Q)}$ and $\frac{1}{\ln(n_{\max})}$ respectively.

\item \textbf{Step 2: Selecting the best pair of smoothness parameters.} For each pair of $\tilde \beta_Q$ and $\tilde \beta_{\max}$ we can construct an estimator $\hat f_{\rm ct}$ as in the CT algorithm. We  select the best smoothness parameters $\beta_Q^*$ and $\beta_{\max}^*$ by minimizing the empirical MSE on the validation data $\D_{Q,2}$. 

\item \textbf{Step 3: Plugging the selected smoothness parameters into the CT algorithm.} Run the CT algorithm with $(\beta_Q^*, \beta_{\max}^*)$ as the smoothness parameters. 
\end{itemize}

The ACT algorithm is summarized in Algorithm \ref{alg: adaptive}. 

\begin{algorithm}[htb] 
\caption{$\A^{\ref{alg: adaptive}}(l)$} \label{alg: adaptive}
\vspace*{0.12 cm}
\KwIn{polynomial degree $l$.}
\vspace*{0.24 cm}
Split $\D_Q$ into $\D_{Q,1}=\{(X_1,Y_1),\dots,(X_{\lfloor \frac{n_Q}{2}\rfloor},Y_{\lfloor \frac{n_Q}{2}\rfloor})\}$ and 
$\D_{Q,2}=\{(X_{\lfloor 
    \frac{n_Q}{2}\rfloor+1},Y_{\lfloor \frac{n_Q}{2}\rfloor+1}),\dots,(X_{ n_Q},Y_{n_Q})\}$.
\\
\vspace*{0.12 cm}
Let $n_{Q,1}=|\D_{Q,1}|$,  $\tilde n_{\max}=\max(n_{Q,1}, n_P)$. Let $\tilde n=\min(\ln(\tilde n_{\max}), {n}_{Q,1})$

Let $S_{\max}=\{\frac{1}{\tilde n},\frac{2}{\tilde n},\dots,\frac{\lfloor(l+1)\cdot \tilde n\rfloor}{\tilde n}\}$. Let $S_Q=\{\frac{1}{\ln(n_{Q,1})},\frac{2}{\ln( n_{Q,1})},\dots,\frac{\lfloor(l+1)\cdot \ln(n_{Q,1})\rfloor}{\ln(n_{Q,1})}\}$
\\
\vspace*{0.12 cm}
\For{$\tilde \beta_Q\in S_Q,\tilde \beta_{\max}\in S_{\max}$}{
Calculate  $\tilde b=\frac{1}{\lfloor \tilde n_{\max} ^{\frac{1}{2\tilde \beta_{\max}+d}}\rfloor}$, $\tilde b_Q=\frac{1}{\lfloor n_{Q,1} ^{\frac{1}{2\tilde \beta_Q+d}}\rfloor}$. \\
\vspace*{0.12 cm}
\If{$n_{Q,1}>n_P$}{
fit local polynomial regression on $\D_{Q,1}$ with bandwidth $\tilde b$, $\hat f_{\rm raw}(\cdot;\tilde \beta_{\max})=\lpr(\cdot;\D_{Q,1},l, \tilde b)$,\\}
\Else{fit local polynomial regression on $\D_{P}$ with bandwidth $\tilde b$, $\hat f_{\rm raw}(\cdot;\tilde \beta_{\max})=\lpr(\cdot;\D_{P},l, \tilde b)$.\\}
\vspace*{0.12 cm}
Fit local polynomial regression on $\D_{Q,1}$ with bandwidth $\tilde b_Q$, and calculate estimates and confidence interval length.\\
$$\hat f_{\rm ref}(\cdot;\tilde \beta_Q)=\lpr(\cdot;\D_{Q,1},l,\tilde b_Q),$$ $$L_{CI}(\tilde \beta_Q)=L_{\rm CI,lpr}(\D_{Q,1},l,\tilde b_Q,\tilde \beta_Q).$$
}
\vspace*{0.12 cm}
Estimate $\hat \psi,\beta_Q^*,\beta_{\max}^*$ by 
$$\hat \psi,\beta_Q^*,\beta_{\max}^*=\argmin_{\psi\in \text{Poly}(\sqrt{\ln( n_Q)}, l), \tilde \beta_Q\in S_Q,\tilde \beta_{\max}\in S_{\max}}\sum_{i=n_{Q,1}+1}^{n_Q}[Y_{i}-\hat f_{\rm ct}(X_i;\psi,\tilde \beta_Q,\tilde \beta_{\max})]^2,$$
where $\hat f_{\rm ct}(x;\psi,\tilde \beta_Q,\tilde \beta_{\max})=\tilde \mu_{\rm ct}(\hat f_{\rm ref}(x;\tilde \beta_Q),\hat f_{\rm raw}(x;\tilde \beta_{\max})+\psi(x),L_{CI}(\tilde \beta_Q)/2)$.\\
\vspace*{0.12 cm}
Truncate the estimate at $n_Q$, 
$$\hat f_{\rm ada}(x)={\rm sgn}(\hat f_{\rm ct}(x;\hat \psi,\beta_Q^*,\beta_{\max}^*))\cdot (|\hat f_{\rm ct}(x;\hat \psi,\beta_Q^*,\beta_{\max}^*)|\wedge \tilde n_Q)$$

\end{algorithm}

Note that in Step 2 we minimize the empirical mean squared error on validation data $\D_{Q,2}$ to select the best polynomial function $\psi$ for each pair of $\tilde \beta_Q$ and $\tilde \beta_{\max}$ and in Step 3 we minimize the same empirical mean squared error on the same validation data to select the best pair of $\tilde \beta_Q$ and $\tilde \beta_{\max}$. Therefore we can combine these two steps in the algorithm table into one step, where we minimize empirical mean squared error on validation data among all choices of $(\tilde \beta_Q,\tilde \beta_{\max})$ and polynomial function $\psi$.

\begin{theorem}[Adaptive upper bound]\label{th: adaptive}
Suppose in Algorithm \ref{alg: adaptive}, $l\geq w(\beta_{\max})\vee T$, then the risk of this algorithm satisfies 
\begin{align*}
&R(\hat f_{\rm ada})=\sup_{(Q,P)\in F(\beta_Q,\beta_P,\epsilon)}  \E (\hat f_{\rm ada}(X)-f(X))^2\\
&\leq C_U^{\rm ada}\cdot\bigg( n_{\max}^{-\frac{2\bmax}{2\bmax+d}}\cdot \ln^4(n_{\max})+ \ln^8(n_Q)\cdot(\epsilon \wedge n_Q^{-\frac{\beta_Q}{2\beta_Q+d}})\cdot n_Q^{-\frac{\beta_Q}{2\beta_Q+d}}+\frac{\ln^4(n_Q)}{n_Q}\bigg),    
\end{align*}

for some constant $C_U^{\rm ada}>0$ not depending on $n_Q,n_{P},\epsilon$.
\end{theorem}

Therefore the data-driven estimator simultaneously achieves the minimax risk, up to a logarithmic factor, for a large collection of parameter spaces.

\begin{remark}{\rm
Theorem \ref{th: adaptive} holds when the intermediate term $L_{\rm CI,lpr}(\D_{Q,1},l,\tilde b_Q,\tilde \beta_Q)$ is equal to $C\cdot \sqrt{\ln(|\D_{Q,1}|)}\cdot (\tilde b_Q^{-\tilde \beta_Q}+\frac{\ln^2(|\D_{Q,1}|)}{\sqrt{|\D_{Q,1}|\cdot {\tilde b_Q}^d}})$ for any constant $C>0$.   In Algorithm \ref{alg: adaptive}, $C$ is taken to be $2$. In practice, it may be beneficial to tune this constant $C$ while using the algorithm. This process can be easily integrated into the algorithm, by tuning $C$ along with the parameters $\tilde \beta_Q$ and $\tilde \beta_{\max}$ on the second half of the dataset.
 }\end{remark}

\section{Simulation} 
\label{section: simulation}
In this section, we evaluate the performance of the ACT algorithm through simulations and compare it to existing methods. The numerical results  further support our theoretical analysis. 

Recall that the minimax risk (\ref{eqn: minimax}) is affected by both the sample size and the bias strength. To demonstrate their impact on the empirical performance, we conduct two series of experiments. In the first series, we fix all other parameters and vary the sample size. In the second series, we fix all other parameters and vary the bias strength.
In all experiments, we set the dimension to $1$, the covariate distributions on both the source and target domains to uniform distribution on $[0,1]$, and the random noise on both domains to normal random variables with zero mean and standard deviation $1/3$. We evaluate the performance of all algorithms using the mean squared error (MSE), which is the expected squared $L_2$ distance between the estimator and the true mean function. The MSE is calculated by averaging 2000 random repeated experiments.

In the first series of experiments, we investigate the influence of sample size by fixing the bias strength and varying the sample size. Specifically we let the mean functions be
\beas
f(x)&=&\sin(10\pi x)+x^{3/2}-0.1x+(0.1-|x-0.5|)_+, \\
g(x)&=&\sin(10\pi x)+x^{3/2}.
\eeas
Therefore $g$ differ from $f$ by a linear function and a small spike with width $0.2$ and height $0.1$. In this case $f\in \H(1,40)$ and $g\in \H(3/2,40)$. The sample size of the target domain is fixed at 200. The sample sizes of the source domain are taken to be $(300,600,1200,2400,4800)$. In this series of experiments, $g$ is smoother than $f$ and $n_P$ is greater than $n_Q$, so $g$ is easier to estimate. 
We compare the performance of the ACT algorithm to that of local polynomial regression using only data from the target domain. The bandwidth for local polynomial regression is determined through a five-fold cross-validation method. By comparing the performance of ACT to local polynomial regression, we are able to gauge the improvement gained through transfer learning with various sample sizes from the target domain.

In the second series of experiments, we investigate the impact of bias strength by fixing the sample size and varying the bias strength. 
Specifically we let the mean functions be 
$$f(x)=\sin(10\pi x)+x^{3/2},$$
$$g(x)=\sin(10\pi x)+x^{3/2}-0.1x+(3-\frac{6}{l_{wid}}|x-0.5|)_+, $$
where $l_{wid}$ is taken to be $0,0.005,0.01,0.015, 0.02$ respectively in each of the five experiments. In this case $g$ is equal to $f$ plus a linear function plus a spike with width $l_{wid}$ and height $3$. Note $\epsilon=l_{wid}*3/2$ in this case. Therefore the bias strengths are $(0,0.00375,0.0075,0.01125,0.015)$. For each of the latter four cases $l_{wid}\in (0.005,0.01,0.015, 0.02)$, $g\in \H(1,\frac{6}{l_{wid}}+40)$ and in the first case where $l_{wid}=0$, $g\in \H(3/2,40)$. 
The sample sizes of the source and target domains are fixed at 200 and 600, respectively. We compare the performance of ACT with local polynomial regression using only the observations from the target domain to study the effects of transfer learning with varying bias strengths. Additionally, comparisons are made with the performance of local polynomial regression using only the observations from the source domain to estimate $g$. These comparisons help illustrate the super-acceleration phenomenon. Both local polynomial regressions are fitted using bandwidths selected through five-fold cross-validation.

\begin{figure}[ht]
    \begin{subfigure}[b]{0.47\textwidth}
      \includegraphics[width=\textwidth]{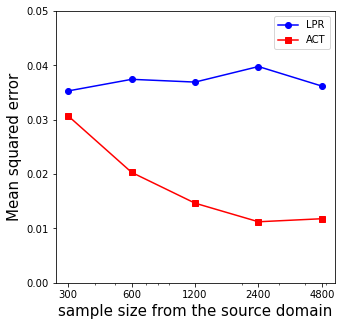}
      \caption{Experiments on different $n_P$}\label{figure: simulation1}
    \end{subfigure}
    \begin{subfigure}[b]{0.48\textwidth}
      \includegraphics[width=\textwidth]{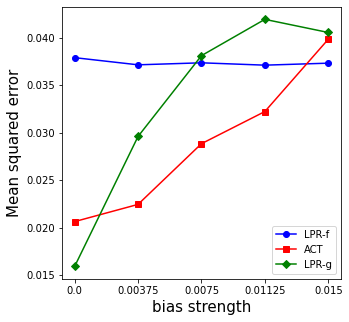}
      \caption{Experiments on different $\epsilon$}\label{figure: simulation2}
    \end{subfigure}
    \caption{MSEs of different regression methods. Blue: MSE of local polynomial regression on the target domain. Red: MSE of ACT on the target domain. Green: MSE of local polynomial regression on the source domain.}
  \end{figure}

Figure \ref{figure: simulation1} presents the results of the first series of experiments, specifically the MSEs of local polynomial regression with cross validation and the ACT algorithm for various sample sizes. As noted, in the first series of experiments, $g$ is smoother and has more corresponding observations, making it easier to estimate. The plot clearly demonstrates the gap in performance between the ACT algorithm and local polynomial regression as predicted by theory. Additionally, the plot indicates that the ACT algorithm's performance improves as the sample size from the source domain increases, however, this improvement seems to level off when $n_P$ is large  ($n_P>2400$).
This is also consistent with the minimax theory. The minimax risk (\ref{eqn: minimax}) in this case is proportional to $$n_P^{-\frac{3}{4}}+(\epsilon\wedge n_Q^{-\frac{1}{3}})\cdot n_Q^{-\frac{1}{3}}+n_Q^{-1},$$
which decreases as $n_P$ grows when $n_P$   is not large and keeps fixed when $n_P$ is large enough such that  $n_P^{-\frac{3}{4}}$ is dominated by the following two terms.

Figure \ref{figure: simulation2} illustrates the simulation results of the second series of experiments. Specifically, it shows the MSEs of local polynomial regression with cross validation for both $f$ and $g$ and ACT algorithm with different bias strength. 
We first compared the ACT algorithm and local polynomial regression for estimating $f$ using observations from the target domain only. The results showed a clear gap in the MSE between the two methods when the bias strength was small enough ($\epsilon<0.01125$). As the bias strength increased, the MSE of ACT grew and eventually became as large as the MSE of local polynomial regression when the bias strength was large enough ($\epsilon>0.015$). These findings are consistent with the theory of minimax risk, which predicts that transfer learning can improve performance when bias strength is small and worsen as it increases. To further illustrate different types of acceleration, we also compared the performance of local polynomial regression for estimating $g$. In the special case of $\epsilon=0$, where $g$ is as smooth as $f$, normal acceleration was observed, as discussed in Section \ref{section: minimax risk}. The results showed that ACT performed worse than estimating $g$ with local polynomial regression but better than estimating $f$ with local polynomial regression. In the general case where $\epsilon>0$, $g$ is rougher than $f$ but has more observations. The theory predicts a super-acceleration phenomenon if the bias strength is small enough. The results showed that when the bias strength was small but nonzero ($0.00375<\epsilon<0.01125$), the ACT algorithm outperformed local polynomial regression for both estimating $f$ and $g$. This validates the theoretical predictions.

\section{Application}
\label{section: application}
In this section, an application of the adaptive estimator is demonstrated using the wine quality data from \cite{cortez2009modeling}. The dataset comprises both red and white wine quality, which share the same features and outcome (wine quality). The aim is to build a regression model that predicts wine quality based on all features. The white wine dataset serves as the source domain and the red wine dataset as the target domain. The objective is to investigate if using the white wine dataset can enhance the prediction of red wine quality.

\begin{figure}[h]
    \includegraphics[width=0.5\textwidth]{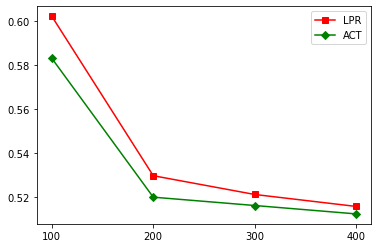}
    \caption{MSEs of different regression methods. Red: MSE of local polynomial regression on the target domain. Green: MSE of ACT.}\label{fig:application}
  \end{figure}

As in Section \ref{section: simulation},  we compare the performance of local polynomial regression applied directly to the red wine dataset with that of our transfer learning algorithm. Both algorithms are based on local polynomial regression, which is suitable for low-dimensional problems. However, the original dataset has 13 features, which are too many for local polynomial regression with the given sample size. To address this, we select the most influential feature, ``alcohol," using feature importance ranking with random forest (\citealp{breiman2001random}) and only use this feature. Both local polynomial regression and the transfer learning algorithm have tuning parameters, so to compare them fairly, we use half of the training samples as the validation dataset to tune the parameters for both algorithms. The degrees of all local polynomials used in both algorithms are set to be $1$. The degree of the polynomial that is used to approximate $f-g$ in ACT algorithm is also set to be $1$.
We let $n_P=4898$ and $n_Q=(100,200,300,400)$. The remaining observations in the target domain serve as the test data to evaluate the performance of both algorithms, which is characterized by the MSE.

Figure \ref{fig:application} shows the MSEs of the two algorithms with different numbers of target sample size. As more observations in the target domain are used, the relative contribution from the source dataset decreases. However, the proposed adaptive estimator consistently outperforms the naive local polynomial regression. This shows that in this application, the performance of the target task can be significantly improved by transfer learning when the source domain has many more observations.

\section{Multiple Source Domains}
\label{section: multiple}

We have so far focused on the single source domain setting. In practical applications, it is common to have data from multiple source domains. In this section, we will expand our analysis to encompass the scenario of utilizing data from multiple source domains, and generalize the procedures and results from the single source domain case to this setting.

We consider the following model  where observations from multiple source distributions $P_1,\dots,P_K$ and one target distribution $Q$ are available. Suppose  there are $n_{P_j}$ observations $\{(X_{1,j}',Y_{1,j}'),\dots,(X_{n_{P_j},j}',Y_{n_{P_j},j}')\}$ from $P_j$ for each $j=1,\dots,K$ and $n_Q$ observations $\{(X_1,Y_1),\dots,(X_{n_Q},Y_{n_Q})\}$ from $Q$. All the observations are independent. Similar to the single-source model, let
\begin{eqnarray*}
Y_{i} &=& f(X_{i})+z_{i}, \quad i=1,\dots,n_Q\\
Y_{i,j}' &=& g_j(X_{i,j}')+z_{i,j}', \quad i=1,\dots,n_{P_j}, \; j=1,\dots,K.
\end{eqnarray*}
where $\{z_{i}\}$ and $\{z_{i,j}'\}$ are i.i.d. zero mean random noises. The parameter space is defined as follows:
\begin{align*}
&F(\beta_Q,\beta_P,\epsilon,u_1,u_2,M,T,\overrightarrow{L_P},L_Q)=\bigg\{(Q,P_1,\dots,P_K):f\in \H(\beta_Q,L_Q),\\
&\quad g_j\in \H(\beta_{P},L_{P_j}), f-g_j\in\Psi(\epsilon,M,T),z_1 |X_1, z_{1,j}' |X_{1,j}'\in G(u_1,u_2),j=1,\dots,K\bigg\},
\end{align*}
where $\overrightarrow{L_P}=(L_{P_1},\dots,L_{P_K})$. This space will be denoted by $F$ for simplicity when there is no confusion. 

We establish the minimax risk in this section,  We also construct a data-driven algorithm, which is an extension of Algorithm \ref{alg: adaptive},  that adaptively achieves the minimax risk up to a logarithmic factor. For reasons of space, the algorithm is given in the Supplementary Material \citep{CaiPu}. 

\begin{theorem}[lower bound]\label{th: lower bound multi}
Let $\bmax=\max(\beta_Q,\beta_P)$ and $n_P=\sum_{j=1}^K n_{P,j}$. Let all assumptions be satisfied, there exists some constant $C>0$ that only depends on $\beta_Q,\beta_P,u_1,u_2,M,K,d,L_P,L_Q$ and not on $n_Q,n_{P,1},\dots,n_{P,K},\epsilon$ such that
 $$R_{\beta_Q,\beta_P,\epsilon}\geq C\cdot\bigg( n_{\max}^{-\frac{2\bmax}{2\bmax+d}}+ (\epsilon \wedge n_Q^{-\frac{\beta_Q}{2\beta_Q+d}})\cdot n_Q^{-\frac{\beta_Q}{2\beta_Q+d}}+\frac{1}{n_Q}\bigg).$$
\end{theorem}

\begin{theorem}[adaptive upper bound]\label{th: adaptive multi}
Suppose in Algorithm \ref{alg: adaptive}, $l\geq w(\beta_{\max})\vee T$, then the risk of this algorithm satisfies
\begin{align*}
&R(n_Q;n_{P})\leq C\cdot\bigg( n_{\max}^{-\frac{2\bmax}{2\bmax+d}}\ln^4(n_{\max})+ \\
&\ln^8(n_Q)\cdot(\epsilon \wedge n_Q^{-\frac{\beta_Q}{2\beta_Q+d}})\cdot n_Q^{-\frac{\beta_Q}{2\beta_Q+d}}+\frac{\ln^4(n_Q)}{n_Q}\bigg),    
\end{align*}
for some constant $C>0$ that only depends on $\beta_Q,\beta_P,u_1,u_2,M,K,d$ and not on $n_Q,n_{P,1},\dots,n_{P,K},\epsilon$.
\end{theorem}

\section{Discussion}
\label{section: discussion}

We studied in the present paper transfer learning for nonparametric regression under the posterior drift model and established the minimax risk, which quantifies when and how much data from the source domains can improve the performance of nonparametric regression in the target domain. A novel, data-driven algorithm is developed and shown to be adaptively minimax optimal, up to a logarithmic factor, over a wide range of parameter spaces.

The minimax risk of this problem exhibits interesting and novel phenomena. The ``auto-smoothing" phenomenon demonstrates that transfer learning can smooth the mean function of the source domain when it is rougher than that of the target domain. The ``super-acceleration" phenomenon shows that even if the task of the source domain is more difficult, it may still be beneficial for the regression task in the target domain in certain cases. Further research in other transfer learning problems could yield similar phenomena.

We use the $L_1$ norm to measure bias strength in this paper, but it is easy to generalize to all $L_p$ norms. This is because $L_1$ norm is smaller than or equal to all $L_p$ norms for $p\geq 1$.  Additionally, polynomial functions are used to approximate the difference between the mean functions of the source and target domains, but it could be interesting to consider other collections of functions in the future. These functions should be easier to estimate than the source and target mean functions, and examples could include infinitely differentiable functions or general \holder  functions with smoothness larger than $\beta_{\max}$.

In this paper, we consider the common support $\Xi$ of the covariates of the source and target domains to be a hypercube of dimension $d$ with edges of length 1, and develop methods and theory for this case.  These results can also be generalized to other types of supports. Specifically, by using linear transformations, our results can be extended to all hypercube-shaped supports. Additionally, we can further generalize our results to more general types of supports by making an assumption on the measure of points not contained in a grid of hypercubes with edge length $\delta$. If this measure is bounded by $O(\delta^\zeta)$ for some $\zeta>0$, our methods and theory can be applied to that support. Examples of supports that satisfy this assumption include all bounded convex sets with $\zeta=1$. Our methods and upper bounds can be adapted to these other supports by considering only the hypercubes contained within them and ignoring the remaining points. The risk of the generalized algorithm is then upper bounded by a constant times the corresponding upper bound for the hypercube support plus $O(\delta^\zeta)$. When $\zeta$ is large enough in relation to the smoothness parameters of the problem, this upper bound matches the lower bound.

\bibliographystyle{apalike}
\bibliography{ref}
\end{document}